\documentclass{article}

\usepackage{PRIMEarxiv}

\usepackage{wrapfig}
\usepackage{pifont}
\usepackage{multirow}
\usepackage{caption}
\usepackage{subcaption}
\usepackage{amsmath} 
\usepackage[utf8]{inputenc} % allow utf-8 input
\usepackage[T1]{fontenc}    % use 8-bit T1 fonts
\usepackage{url}            % simple URL typesetting
\usepackage{booktabs}       % professional-quality tables
\usepackage{amsfonts}       % blackboard math symbols
\usepackage{nicefrac}       % compact symbols for 1/2, etc.
\usepackage{microtype}      % microtypography
\usepackage{lipsum}
\usepackage{fancyhdr}       % header
\usepackage{graphicx}       % graphics
\graphicspath{{media/}}     % organize your images and other figures under media/ folder

\usepackage[pagebackref,breaklinks,colorlinks]{hyperref}

\title{Dealing with Subject Similarity in \\ Differential Morphing Attack Detection
}

\author{
  Nicolò Di Domenico, Guido Borghi, Annalisa Franco, Davide Maltoni \\
  Department of Computer Science and Engineering \\
  University of Bologna \\
  \texttt{\{name.surname\}@unibo.it} \\
}

\begin{document}
\maketitle

\begin{abstract}
The advent of morphing attacks has posed significant security concerns for automated Face Recognition systems, raising the pressing need for robust and effective Morphing Attack Detection (MAD) methods able to effectively address this issue.
In this paper, we focus on Differential MAD (D-MAD), where a trusted live capture, usually representing the criminal, is compared with the document image to classify it as morphed or bona fide. 
We show these approaches based on identity features are effective when the morphed image and the live one are sufficiently diverse; unfortunately, the effectiveness is significantly reduced when the same approaches are applied to look-alike subjects or in all those cases when the similarity between the two compared images is high (e.g. comparison between the morphed image and the accomplice).
Therefore, in this paper, we propose ACIdA, a modular D-MAD system, consisting of a module for the attempt type classification, and two modules for the identity and artifacts analysis on input images. 
Successfully addressing this task would allow broadening the D-MAD applications including, for instance, the document enrollment stage, which currently relies entirely on human evaluation, thus limiting the possibility of releasing ID documents with manipulated images, as well as the automated gates to detect both accomplices and criminals.
An extensive cross-dataset experimental evaluation conducted on the introduced scenario shows that ACIdA achieves state-of-the-art results, outperforming literature competitors, while maintaining good performance in traditional D-MAD benchmarks.
\end{abstract}

\section{Introduction} \label{sec:introduction}
The recently introduced Morphing Attack~\cite{matteo2014magic} poses a significant security threat in face verification-based applications, systems usually exploited for instance in the Automatic Border Control (ABC) gates in international airports~\cite{scherhag2017biometric}.
Indeed, through this attack, it is possible to obtain a regular and legal document that presents a morphed photo, \textit{i.e.} a hybrid face image that hosts two different identities, and that can be shared between two subjects. In this way, a \textit{criminal} can bypass official controls using the identity of an \textit{accomplice} without any criminal record. 

In this context, the development of effective Morphing Attack Detection (MAD) systems~\cite{raja2020morphing}, \textit{i.e.} methods able to automatically detect the presence of the morphing procedure on input images, is strongly demanded by private and public institutions. 
Generally, two families of MAD detectors are investigated in the literature~\cite{borghi2022incremental}: Single-Image MAD (S-MAD) methods, which usually rely on the detection of the presence of visible or invisible morphing-related artifacts in the single input image, and Differential MAD methods (D-MAD), in which recent SoA methods~\cite{scherhag2020deep,kessler2023towards,borghi2021double} commonly compare the identity of the two facial images -- the document and the trusted live acquisition image -- received as input.

A successful morphing attack requires fooling the human examiner at the document enrollment stage, typically presenting a morphed photo very similar to the document applicant, and fooling with the same morphed image the automatic face verification system, \textit{e.g.} to pass the security check at ABC gates. 
Therefore, the morphed image is often created with a stronger presence of the accomplice, so that the human examiner is less inclined to notice noticeable differences. 
Usually, this critical aspect is controlled through the morphing factor parameter, typically set to $0.3$~\cite{ferrara2019decoupling}, thus conferring a preference for the resemblance of the morphed image to that of the accomplice.

We observe that, unfortunately, most of the D-MAD approaches available in the literature have been developed thinking of a practical application at the verification stage, \textit{i.e.} when the criminal subject tries to use the passport for instance at ABC gates. 
Therefore, the large majority of existing benchmarks only consider the comparison between the document image and the criminal or the bona fide subject.

\begin{figure}[t!]
% \begin{wrapfigure}{l}{0.45\textwidth}
    \centering
    \includegraphics[width=0.5\columnwidth]{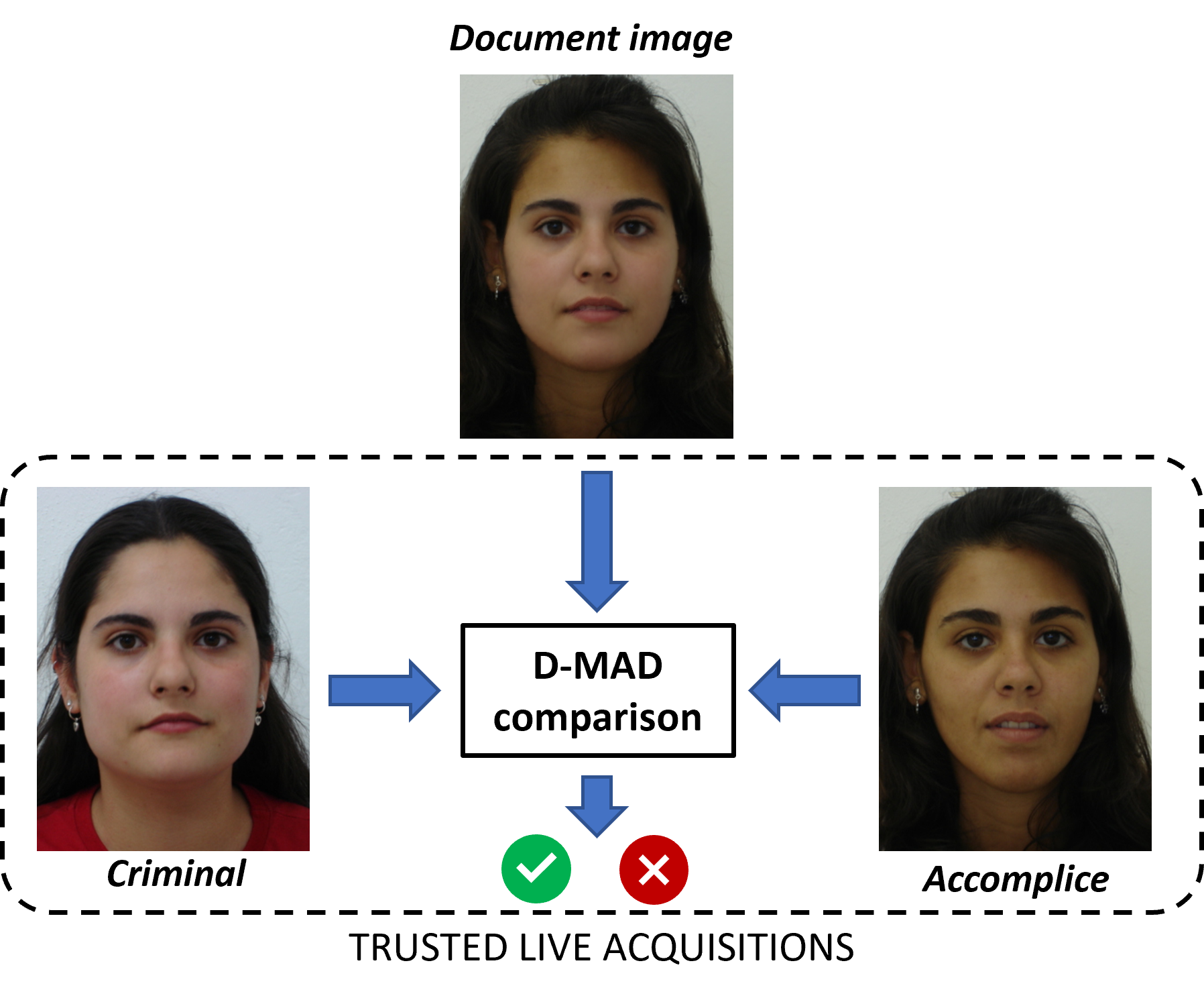}
    \caption{In traditional D-MAD benchmarks, usually the document image is compared only with bona fide and criminal live photos. 
    Therefore, in this paper, we introduce a new challenging scenario in which the document image is compared with both the criminal and the accomplice. From a practical point of view, this scenario broadens the applications of D-MAD systems.}
    \label{fig:initial}
\end{figure}
% \end{wrapfigure}

Differently, in this paper, we consider a new scenario in which D-MAD is applied to both criminal and accomplice verification attempts, as shown in Figure \ref{fig:initial}.
We observe that successfully tackling this benchmark would broaden the applicative context of D-MAD methods, that can be exploited also at the enrollment stage in order to avoid a morphed image being included in the document.
Besides, this scenario would allow the use of MAD methods in ABC gate to identify not only the criminal but also the accomplice, who has stained themselves with a criminal and punishable action during the document issuing procedure. It is also worth considering that, since in the morphing attack the accomplice applies for a valid passport, he will have to use that passport for several years if he needs to travel, thus making of interest the detection of the accomplice's attempt too.
Although the use of the morphed image by an accomplice is maybe less pressing, the development of such D-MAD technologies could still be of interest in the future, even as a mere precautionary measure.
We also note that current identity-based D-MAD methods are negatively influenced by the high similarity between the document image and the accomplice, as detailed and experimentally analyzed in Section~\ref{sec:mad_issues}, and then this scenario represents an interesting and challenging research field, that needs to be addressed in future MAD-related work.

Therefore, we propose ACIdA, a D-MAD method that specifically tackles this challenging novel scenario, based on three modules that focus on different aspects of the problem: the Attempt Classification (AC) module, responsible for the classification of the identity verification attempt in three different classes (\textit{i.e.} criminal, accomplice or bona fide attempt); the Identity-Aritfact module (IdA), based on a combination of artifacts detection and identity analysis; the Identity module (Id) based only on identity comparison.

The new scenario considered in this paper has been first introduced in ~\cite{di2023combining} where we showed that combining identity features and artifact analysis can help to better deal with high subject similarity. This paper further extends that initial work, including the following original contributions:

\begin{figure*}[th!]
    \centering
    \includegraphics[width=0.95\linewidth]{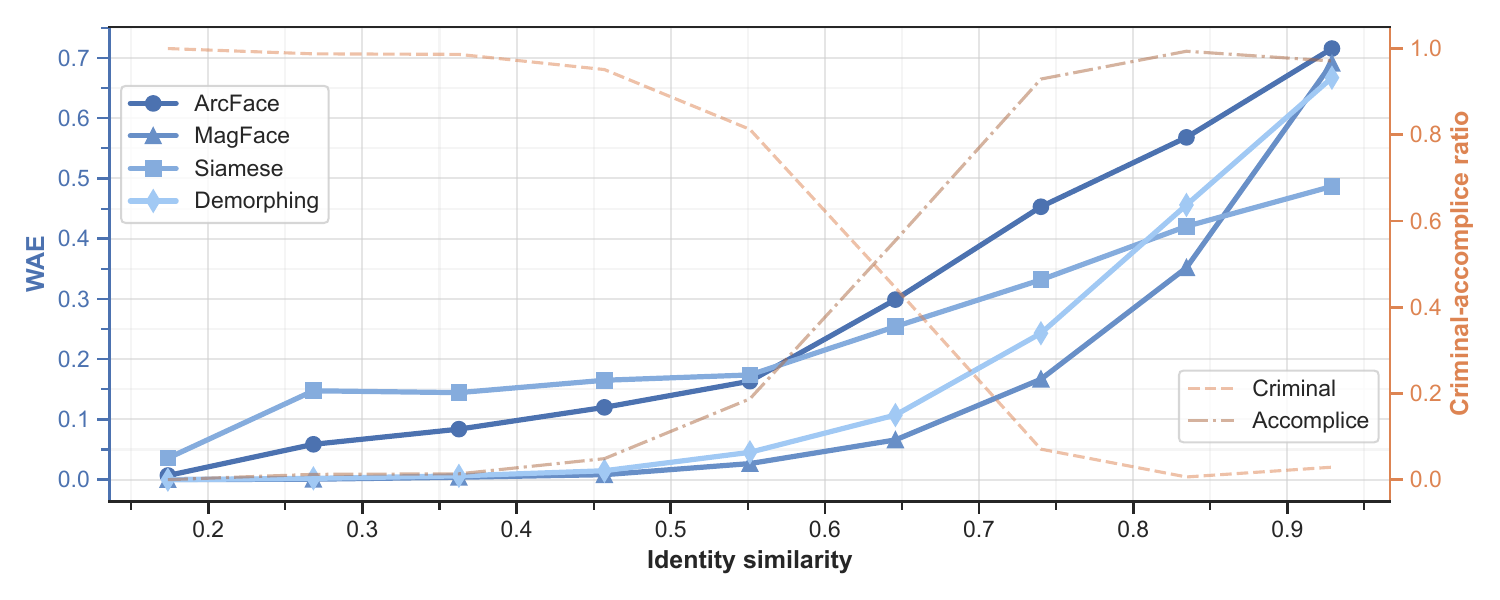}
    \caption{Performance of literature D-MAD methods across different identity similarity scores (\textit{x}-axis) between input subjects. Values close to $1$ indicate a high similarity. MAD performance is expressed as the weighted average of the error-based metrics commonly used in the MAD task, and here referred to as WAE (see Sect.~\ref{sec:metrics}). As reported, all methods are negatively influenced by the increasing identity similarity, which corresponds to the increase in the accomplice's presence in input images instead of the criminal. 
    The criminal-accomplice ratio is computed as the number of pairs with the criminal and the accomplice over the total.
    Both percentages are represented with dotted lines. 
    The analyzed D-MAD methods are as follows: ArcFace~\cite{scherhag2020deep} (circle), MagFace~\cite{kessler2023towards} (triangle), Demorphing~\cite{ferrara2017face} (rhombus), and Siamese~\cite{borghi2021double} (square).}
    \label{fig:mad_vs_id}
\end{figure*}

\begin{itemize}

    \item We investigate a D-MAD scenario in which document images are compared with both accomplice and criminal contributing subjects and the morphing process in not necessarily symmetric (\textit{i.e.} it does not include an equal contribution of the two subjects). From a practical standpoint, this approach broadens the scope of application of current D-MAD methods to scenarios, such as the enrollment one, that have been relatively unexplored until now. We also highlight, from a quantitative point of view, the lack in terms of accuracy of identity-based D-MAD methods currently available in the literature.

    \item We propose ACIdA, a deep learning-based D-MAD method expressively conceived for the new scenario. Specifically, with respect to \cite{di2023combining}, we introduce a new module to provide the attempt classification, and the other two modules focus on identity and artifact features. All these modules cooperate together to output the final score through a weighted prediction.
    For the sake of reproducibility, we publicly release the code, the list of subject pairs used for morphing generation for training and testing sets, and the implementation details of the proposed method\footnote{\url{https://github.com/ndido98/acida}}.
    
    \item We perform an extensive and cross-dataset evaluation, highlighting several experimental aspects of the proposed method. Moreover, we show the generalization capability of the proposed method in a different D-MAD scenario through the FVC-onGoing~\cite{fvc_ongoing} platform.

\end{itemize}

\subsection{Issues with existing D-MAD methods} \label{sec:mad_issues}
Differential Morping Attack Detection (D-MAD) methods, also known as two-image- or pair-based methods, involve the utilization of a pair of face images as input.
The aim of these methods is to determine whether the document image has been morphed or not. 
The second image, which is a trusted live capture, can serve as a helpful reference during the task, such as for analyzing identity, texture, or color inconsistencies when compared with the first one. 

In practice, these two images can be obtained during passport issuance, where the first image is the provided photo and the second image is captured in real-time. Similarly, during controls at ABC gates, the live image is acquired through automated face verification procedures, while the probe image is retrieved from the electronic Machine Readable Travel Document (eMRTD).

Analyzing the results obtained on two important sequestered datasets, \textit{i.e.} NIST FATE MORPH~\cite{cit:nist} and FVC-onGoing~\cite{dorizzi2009fingerprint}, it can be noted that the most promising D-MAD algorithms are based on the comparison of the subject identity through pre-trained deep learning architectures. 
We observe that these approaches can be limited in accuracy with look-alike subjects and, in particular, when morphed images are compared with the accomplice.
These hypotheses are preliminarily confirmed in our experiments, the results of which are reported in Figure~\ref{fig:mad_vs_id}.
In particular, we test the performance of various literature D-MAD systems across several image pairs with increasing similarity between the subjects, \textit{i.e.} the document image and the live acquisition.
Identity similarity is obtained by computing the cosine similarity of identity features extracted through the method described in~\cite{meng2021magface}. For a comprehensive analysis, we also computed the percentage of criminal and accomplice pairs over the total number of attempts as a function of similarity.
In order to summarize the MAD performance, results are reported through the WAE metric (detailed in Sect.~\ref{sec:metrics}), \textit{i.e.} the average of error-based metrics commonly used in the MAD task.
As shown, all reported methods suffer the increasing similarity (values close to $1$ on \textit{x}-axis), which corresponds to an increasing number of verification attempts involving the accomplice subject instead of the criminal one (represented with dotted lines). 
The trend of the method~\cite{borghi2021double} differs from other algorithms because, despite slightly lower performance for low similarity values, it exhibits greater robustness as the similarity value increases. One possible explanation derives from the fact that this method utilizes both identity-related features and features associated with the presence of artifacts to compute the final morphing score (see Sect.~\ref{sec:related_work} for more details).

\begin{figure}[t!]
     \centering
     \begin{subfigure}[b]{0.18\columnwidth}
         \centering
         \includegraphics[width=\textwidth]{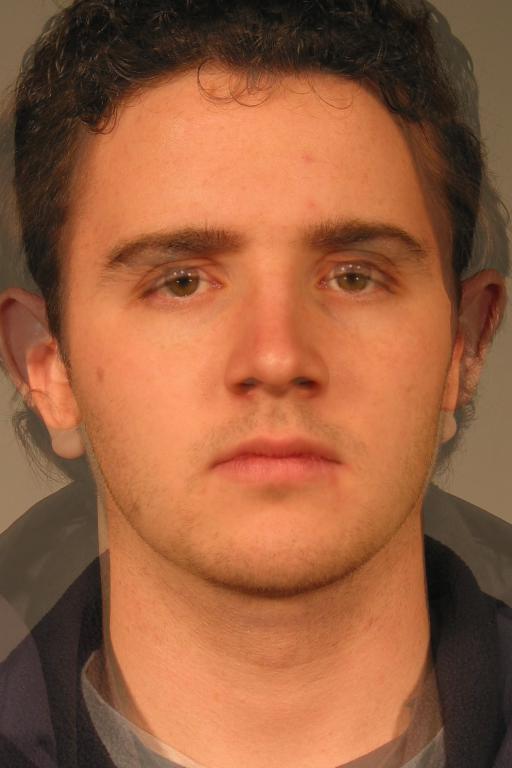}
         % \caption{}
         % \label{fig:y equals x}
     \end{subfigure}
     \medspace
     \medspace
     \begin{subfigure}[b]{0.18\columnwidth}
         \centering
         \includegraphics[width=\textwidth]{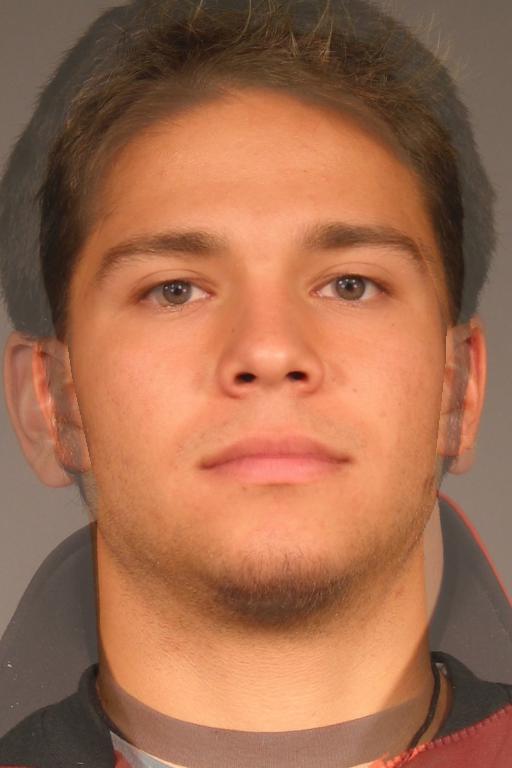}
         % \caption{}
         % \label{fig:three sin x}
     \end{subfigure}
     \medspace
     \medspace
     \begin{subfigure}[b]{0.18\columnwidth}
         \centering
         \includegraphics[width=\textwidth]{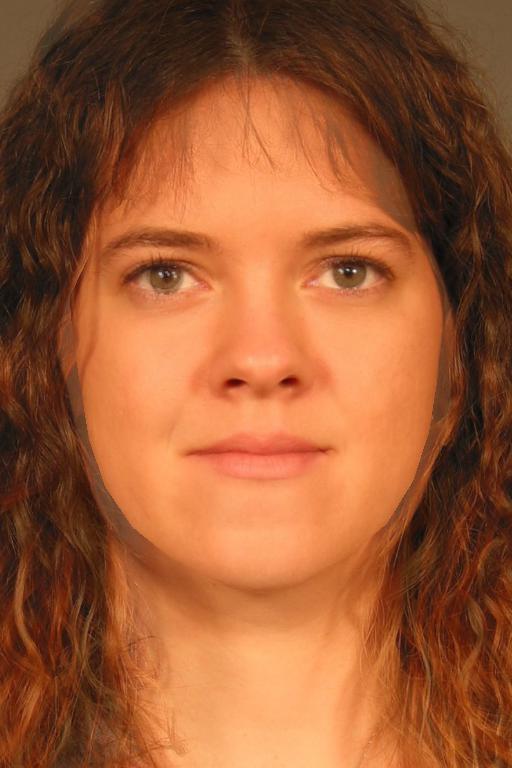}
         % \caption{}
         % \label{fig:five over x}
     \end{subfigure}
        \caption{Sample of morphed images that successfully fool the identity-based D-MAD system~\cite{scherhag2020deep} but that exhibit visible artifacts related to the morphing procedure. Hence, the intuition to exploit also artifact detection techniques to improve the performance in the D-MAD task, as described in Section~\ref{sec:mad_issues}.}
        \label{fig:bad_mad}
\end{figure}

Furthermore, an additional limitation of current D-MAD methods lies in the robustness of pre-trained networks on large and diverse datasets for the face recognition task, which makes these methods potentially insensitive to clearly visible artifacts in morphed images. An example of this is depicted in Figure~\ref{fig:bad_mad}, in which the SoA D-MAD method described in~\cite{scherhag2020deep} wrongly predicts these images as bona fide, even though visible artifacts are present.
Hence, and also taking into account also the previous considerations, the intuition to exploit also artifact detection in our D-MAD method, as detailed in Section \ref{sec:Ida_module}.

\section{Related Work} \label{sec:related_work}
A variety of D-MAD methods based on hand-crafted features, such as BSIF~\cite{kannala2012bsif}, LBP~\cite{ojala1994performance}, and HOG~\cite{dalal2005histograms}, and Machine Learning classifiers have been proposed in the literature, achieving partially satisfactory results~\cite{fvc_ongoing,cit:nist}.
In particular, the method described in~\cite{scherhag2018towards} proposes to compute a histogram of LBP for both images and to combine them via subtraction. Then, the resulting $256$-dimensional feature vector is finally used as input to train a Support Vector Machine (SVM), that is ultimately responsible for emitting the morphing score.
A small portion of methods relies on comparisons between specific elements of the face, such as the work described in \cite{scherhag2018detecting} that is based on the Euclidean and angle distance between the facial landmarks of the probe and live images. However, the final performance is strongly limited by the abilities of the facial landmark predictor, and then this category of methods is excluded from our analysis.

However, one of the most effective and accurate D-MAD approaches is presented in the recent work of Scherhag \textit{et al.} ~\cite{scherhag2020deep}, in which authors propose a Convolutional Neural Network (CNN) architecture, specifically a ResNet50~\cite{he2016deep}, trained with an angular margin loss referred to as ArcFace~\cite{deng2019arcface}, originally designed for the Face Recognition task. 
This architecture is utilized to extract feature embeddings from the input images. According to the authors, the network is pretrained and no additional training procedures are performed specifically for the morphing detection task: this ensures that the deep learning model does not suffer from overfitting with training datasets limited in size and variety.
Finally, the extracted features are then subtracted and fed into an SVM for the final classification process.
A recent evolution of this system has been proposed in~\cite{kessler2023towards}, in which the backbone is a ResNet trained through the MagFace~\cite{meng2021magface} loss function, an adaptive mechanism to learn a well-structured within-class feature distribution relying on the magnitude of vectors that have achieved SoA performance on the Face Recognition task.

A similar approach is described in~\cite{ferrara2017face}, where the reverse morphing procedure (referred to as ``demorphing'') is applied to the input images in order to reveal the real identity hidden in the morphed image. This method works in combination with Commercial-On-The-Shelf (COTS) face verification systems and is interesting since it does not require any training procedure. In this case, the main issue is represented by the fact that the morphing process rarely is a simple linear combination as assumed by the authors. Moreover, the entire process is based on the accuracy of the estimation of the position of facial landmarks and even small localization errors could negatively influence the effectiveness of the whole pipeline.
Similar approaches based on the Generative Adversarial Network (GAN) paradigms have been proposed~\cite{peng2019fd} in order to restore the accomplice's facial image hidden in the morphed one, unfortunately with limited generalization capabilities.

In~\cite{chaudhary2021differential}, the authors propose to exploit the undecimated 2D Discrete Wavelet Transform information to feed a Siamese neural network, that accentuates the disparity between real and morphed images. The results reported in the original paper are interesting, even though the lack of implementation details (at the time of writing, the method is under patenting) limits the reproducibility.

The work described in~\cite{borghi2021double}, proposes another deep learning-based Siamese network in order to focus not only on features related to the identity of input subjects but also on the presence of artifacts left by the morphing procedure. Two different branches process the same input image and their output is finally combined through a fully connected layer to obtain the final morphing score. 

Recently, various works have tackled the D-MAD task, using a variety of different techniques, such as the fusion of the output of different backbones~\cite{medvedev2023fused, shiqerukaj2022fusion} or the feature-wise supervision on fine-grained classification~\cite{qin2022face}. The work of Singh et al.~\cite{singh2022reliable} describes a method based on the fusion of several deep features computed from six different CNNs, trained on the ImageNet dataset, and merged through a spherical interpolation, referred as SLERP. Differently from ours, this method is not based on identity features, and it is specifically conceived for the on-the-fly D-MAD task with different camera resolutions and acquisition distances.

\section{Proposed Method}
The underlying idea of the development of our method is to process every possible attempt -- criminal, accomplice, and bona fide -- with a specific approach based on different features. 
Our insight is that for criminal and bona fide verification attempts, identity-related features are very effective, as confirmed by the good results achieved by~\cite{scherhag2020deep}. 
Differently, for accomplice verification attempts, due to the greater similarity between subjects, the discriminative power of identity features is reduced (see Fig.~\ref{fig:mad_vs_id}) and we believe the combination with artifact analysis can improve the MAD performance. 
To accomplish this paradigm, an initial selector of the attempt type is needed: then, we build a classifier that outputs the probability of the attempt type used in the final MAD score computation.
The effectiveness of the use of different modules is experimentally confirmed by the ablation study reported in Section~\ref{sec:results}.

\begin{figure*}[t!]
    \centering
    \includegraphics[width=0.95\linewidth]{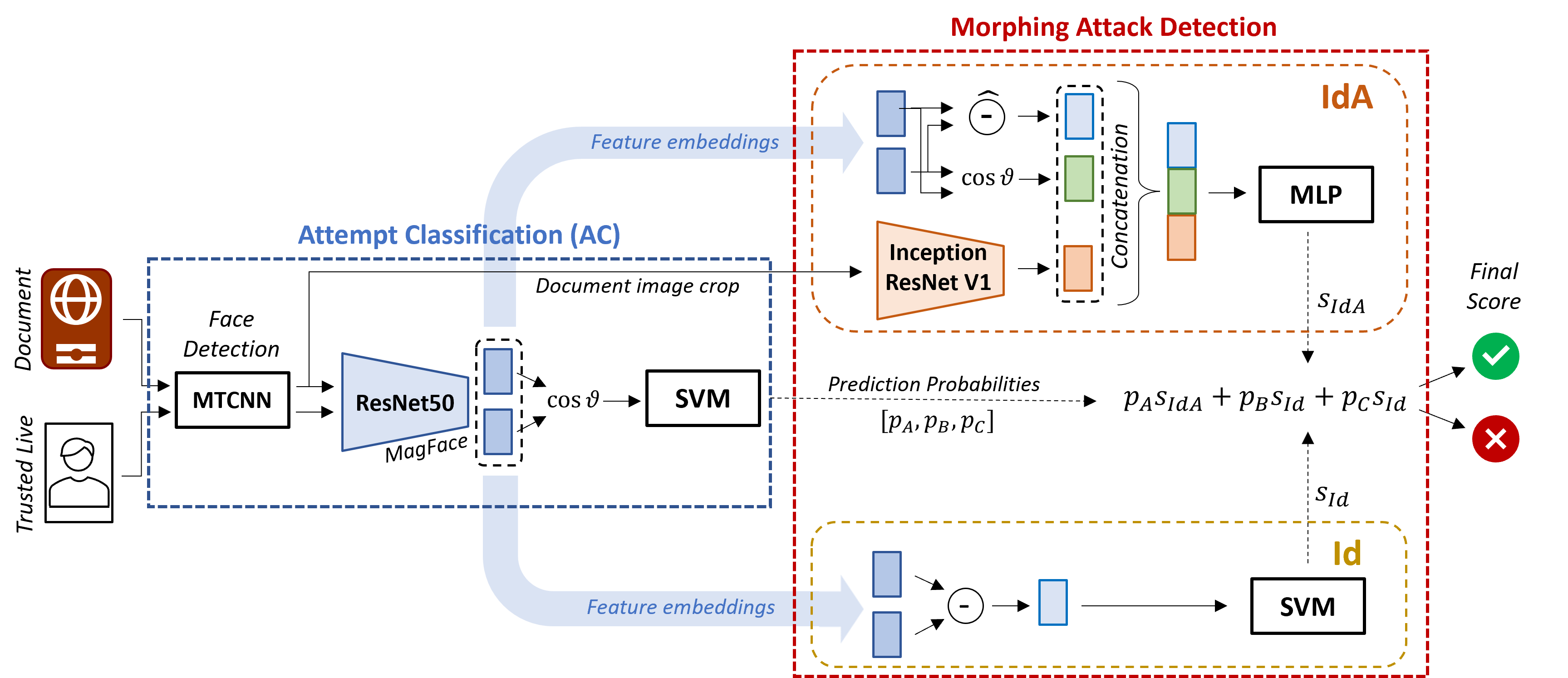}
    \caption{General overview of the proposed system, named ACIdA. As shown, the method is composed of three different modules: the Attempt Classification (AC) module, responsible to determine if the document image is compared with the criminal or the accomplice trust live image (see Sect.~\ref{sec:ac_module}); the Identity (Id) module, an identity comparison-based MAD system (see Sect.~\ref{sec:id_module}) and the Identity-Artifact (IdA) module, that integrates both information about identity and artifact detection (see Sect.~\ref{sec:Ida_module}). 
    Finally, the score of these two MAD modules is combined through a weighted sum to produce the final output of the system, \textit{i.e.} predicting if the input document image is morphed or bona fide.}
    \label{fig:general}
\end{figure*}

A general overview of the proposed method is depicted in Figure~\ref{fig:general}.
As shown, the framework is divided into three main modules: the Attempt Classification (AC) module, assigned to the classification of the pair attempt provided in input, the Identity-Artifact (IdA) block, specialized in the detection of morphed images combining identity and artifact information, and the Identity block (Id), that relies only on identity analysis to detect morphing attacks.

The final output score is obtained through a weighted sum, formally defined as:
\begin{equation}
    S = p_A \cdot S_{IdA} + p_B \cdot S_{Id} + p_C \cdot S_{Id}
    \label{eq:weighted}
\end{equation}
where $p_A, p_B, p_C$ are the probabilities that the document image is compared with the accomplice, bona fide, and criminal subjects, respectively; these values are output by the SVM classifiers of the AC module. $S_{IdA}, S_{Id}$ are the outputs produced by the classifiers of the IdA and Id modules. 
With this procedure, the final output is a weighted sum of all contributions produced by the different modules, described in the following. 

\subsection{Attempt Classification module} \label{sec:ac_module}
The Attempt Classification module is responsible for preparing the input for all the other modules and for the pair typology classification itself, \textit{i.e.} the prediction if the subject depicted in the live acquisition (attempt) is the criminal or the accomplice (in case of morphed document image), or the same subject (bona fide). 
The input is represented by the probe image, \textit{i.e.} the one contained in the document, and the trusted live acquisition. Firstly, these images are fed into the MTCNN face detector~\cite{zhang2016joint} that, as the name suggests, crops the face excluding the large part of the background, preparing the input images for the feature extraction procedure. 
The resulting crops are then used as input to extract features through a backbone that outputs two feature embeddings of size $512$. 
We employ a frozen iResNet~\cite{duta2021improved} architecture as backbone, trained using the magnitude and angular MagFace loss~\cite{meng2021magface}, originally conceived for the face recognition task. This training process has been conducted on vast-scale datasets, namely MS-Celeb-1M~\cite{guo2016ms} and VGG-Face2~\cite{cao2018vggface2}, which consist of trillions of face pairs. 
The two produced embeddings are combined through the cosine similarity and are finally used as input for an SVM classifier that outputs $3$ possible different classes: ``bona fide'', ``accomplice'' and ``criminal'' along with their probabilities used for the final classification, as previously detailed.

\subsection{Identity-Artifact (IdA) module} \label{sec:Ida_module}
As the name suggests, the idea behind this module is to combine information belonging to two different tasks, \textit{i.e.} face recognition and artifact detection, following the considerations reported in Section \ref{sec:mad_issues}.

This module receives as input the extracted feature embeddings and the document images crops.
The feature embeddings are then combined in two different ways: in the first, a subtraction followed by a min-max normalization (that rescales each component in the range $[0, 1]$) is exploited, while in the second the cosine similarity is used.
Since these embeddings are produced starting from a backbone trained for the Face Recognition task, we assume they contain information about the subjects' identities. 

The document image crops are fed into an Inception-ResNet V1~\cite{szegedy2015going} architecture, starting from the weights obtained with the VGG-Face2~\cite{cao2018vggface2} datasets. An additional fine-tuning procedure is conducted on ICAO-compliant and JPEG-compressed images. A $512$-dimensional feature embedding is obtained removing the final last fully connected layer of the adopted architecture.

These three outputs are finally combined through a concatenation, obtaining a $1025$-dimensional feature vector, used as input to a Multi-Layer Perceptron (MLP) architecture exploited as a classifier that produces in output a score in the range $\left[0, 1\right]$.

\subsection{Identity (Id) module} \label{sec:id_module}
For the identity module, we draw inspiration from the solution presented in~\cite{scherhag2020deep}, which can be considered the current state-of-the-art D-MAD method, as evidenced by the results published on the FVC-onGoing platform~\cite{dorizzi2009fingerprint}.

We employ an iResNet~\cite{duta2021improved} network that has been trained for the purpose of face recognition using the MagFace loss~\cite{meng2021magface}. As the input consists of two images, this module generates two distinct feature embeddings of size $512$. These embeddings are then combined through subtraction, resulting in a single final feature vector of the same size that is fed into an SVM classifier, trained to output a probability in the $\left[0, 1\right]$ range that represents whether the probe image is the result of a morphing process. 
The authors demonstrate that the MagFace loss function yields robust embeddings by maximizing the geodesic distance between different identities. Consequently, the produced embeddings exclusively contain information pertaining to the input face identity.

\subsection{Proposed Scenario}
As previously mentioned, in the proposed scenario the document images are compared with bona fide, criminal, and accomplice subjects.
In particular, for the sake of clarity, the experimental results are categorized into three distinct sub-benchmarks based on the trusted live image's identity. 

\begin{figure}[th!]
     \centering
     \begin{subfigure}[b]{0.18\columnwidth}
         \centering
         \includegraphics[width=\textwidth]{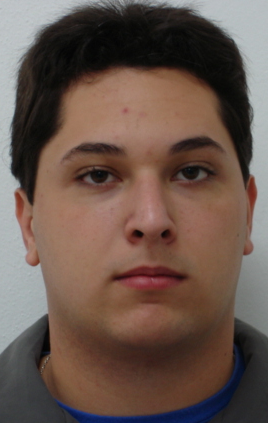}
         \caption{Subject 1}
         \label{fig:acc}
     \end{subfigure}
     \medspace
     \medspace
     \begin{subfigure}[b]{0.18\columnwidth}
         \centering
         \includegraphics[width=\textwidth]{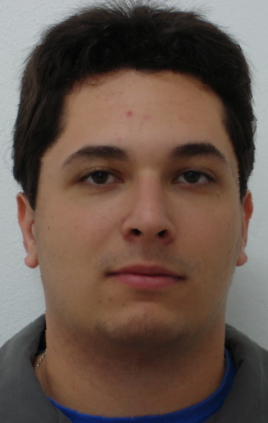}
         \caption{Morphed}
         \label{fig:morphed}
     \end{subfigure}
     \medspace
     \medspace
     \begin{subfigure}[b]{0.18\columnwidth}
         \centering
         \includegraphics[width=\textwidth]{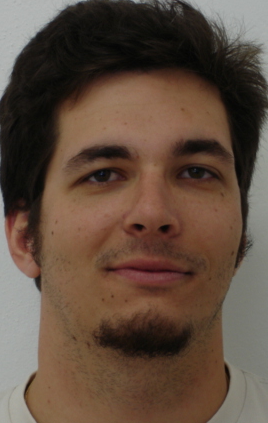}
         \caption{Subject 2}
         \label{fig:criminal}
     \end{subfigure}
        \caption{In a D-MAD approach, the morphed image (Fig.~\ref{fig:morphed}) can be compared with the criminal (Fig.~\ref{fig:criminal}) or with the accomplice (Fig.~\ref{fig:acc}), or both, as in the introduced scenario.}
        \label{fig:sample_couple}
\end{figure}

These benchmarks include: 
\begin{itemize}
    
    \item \textbf{Accomplice}: this benchmark is built with morphed attempts where the live image is from the accomplice as well as bona fide attempt. An example of the input pair of this benchmark consists in Figure~\ref{fig:acc}~and~\ref{fig:morphed}.
    
    \item  \textbf{Criminal}: conversely, this scenario encompasses authentic and morphed attempts where the live image belongs to the criminal subject (\textit{e.g.} Figure~\ref{fig:criminal}~and~\ref{fig:morphed}). This is the common case usually investigated in the MAD literature. 
    
    \item \textbf{Both}: in this case, this scenario is obtained through a union of the previous ones. This scenario is useful to understand the generalization capabilities of the MAD method, which is not aware of the pair type received in input.
\end{itemize}

In the subsequent analysis, particular emphasis is placed on the performance achieved in the ``accomplice'' scenario. 
As previously mentioned and experimentally validated (see Fig.~\ref{fig:mad_vs_id}), due to the greater resemblance between the subjects depicted in both images, this scenario is generally regarded as more challenging for D-MAD methods based on identity analysis.

Finally, it is important to note that the ``criminal'' case represents the traditional benchmark for testing D-MAD methods, so these results can be used as a reference for previous work.

\subsection{Datasets} \label{sec:datasets}
\noindent\textbf{Progressive Morphing Database (PMDB)}~\cite{ferrara2017face}: a total of $1108$ morphed images are obtained by utilizing three commonly used datasets in the MAD field, namely AR~\cite{martinez1998ar}, FRGC~\cite{phillips2005overview}, and Color Feret~\cite{phillips1998feret}. 
These images have been generated using a publicly available morphing algorithm outlined in a previous study~\cite{ferrara2017face}. The morphing process involved a cohort of $280$ individuals, consisting of $134$ males and $146$ females. It is noteworthy that these morphed images have not been subjected to manual retouching procedures to enhance their visual quality, thus potentially exhibiting artifacts such as blurred areas or ghost effects. However, it is important to mention that the background replacement performed by the morphing is artifact-free.

\noindent\textbf{Idiap Morph}~\cite{sarkar2022gan} is a collection of multiple datasets publicly accessible, comprising five subsets generated using different morphing algorithms. In our analysis, we focus on OpenCV~\cite{opencv_morph}, FaceMorpher~\cite{facemorpher}, and StyleGAN~\cite{karras2020analyzing}.
These algorithms utilize face images from the Feret, FRGC, and Face Research Lab London Set~\cite{debruine2017face} datasets as input data.
The overall visual quality of the morphed images created with OpenCV and FaceMorpher is negatively affected by the presence of various artifacts present in both the background and the foreground of the images. Differently, morphed faces generated through StyleGAN exhibit fewer visible visual artifacts, but common textures associated with GANs are still discernible.

\noindent\textbf{MorphDB}~\cite{ferrara2017face}. This dataset is constructed using images sourced from the Color Feret~\cite{phillips1998feret} and FRGC~\cite{phillips2005overview} datasets. It comprises 100 morphed images generated through the Sqirlz Morph 2.1 algorithm. This dataset offers valuable material as all morphed images have undergone manual retouching, resulting in good final visual quality.

\noindent\textbf{FEI Morph.} This dataset in question is produced using images sourced from the FEI Face Database~\cite{thomaz2010new}, which consists of $200$ subjects evenly divided between males and females. The faces within the database predominantly represent individuals aged between $19$ and $40$ years old, showcasing distinct appearances, hairstyles, and accessories. This dataset comprises a total of $6000$ morphed images, generated through the utilization of three different morphing algorithms: FaceFusion~\cite{facefusion}, UTW~\cite{raja2020morphing}, and NTNU~\cite{raja2020morphing}. These algorithms employ two varying morphing factors, specifically $0.3$ and $0.5$.
The need to introduce this new dataset arises from the necessity to faithfully replicate the new scenario introduced, in which, specifically, the morphed image appears particularly similar to the accomplice and the goal is to detect morphing even when the live acquisition comes from the most similar subject.

\subsection{Experimental Protocol}
In all our experiments, we conduct a cross-dataset evaluation to assess the effectiveness of our method. The training and validation of the proposed approach are performed on PMDB, MorphDB, and Idiap Morph datasets, while the proposed MAD method is tested on the FEI Morph dataset.
We observe that the testing FEI Morph dataset is completely disjoint from the training ones: it has no common elements with the training datasets, neither in terms of contributing subjects (for the morphing generation) nor in terms of the morphing algorithms.
It is important to note that the test on the FEI Morph dataset is fully reproducible for future comparisons since the FEI Face Database~\cite{thomaz2010new} used to select the input images is publicly available as well as the morphing algorithms used to generate the morphed images; in addition, the list of image pairs used for morphing generation and the index of test attempts will be released.
A cross-dataset evaluation procedure is relevant due to the scarcity of publicly available datasets with diverse samples for each subject and morphing algorithms. Besides, privacy issues play a crucial role in hindering the public release of such datasets. 

\subsection{Metrics} \label{sec:metrics}
In the evaluation and comparison of MAD systems, various metrics are typically employed to assess their performance~\cite{raja2020morphing}. 
Two commonly used metrics are referred to as Bona Fide Presentation Classification Error Rate (BPCER) and the Attack Presentation Classification Error Rate (APCER).
The BPCER measures the proportion of bona fide images that are incorrectly classified while the APCER represents the proportion of morphed images that are erroneously labeled as bona fide.

The formulations of these metrics are as follows:
\begin{equation}
    \textnormal{BPCER} (\tau) = \frac{1}{N} \sum_{i=1}^{N} H(b_i - \tau)
\end{equation}
\begin{equation}
    \textnormal{APCER} (\tau) = 1 - \left[ \frac{1}{M} \sum_{i=1}^{M} H(m_i - \tau) \right]
\end{equation}
In both definitions, $\tau$ represents the score threshold at which the detection scores for bona fide and morphed images ($b_i,m_i$) are compared. The function $H(x)$ is defined as a step function, which returns $1$ if $x$ is greater than $0$ and $0$ otherwise.
The BPCER is typically evaluated with respect to a specified APCER value, here referred to as \textbf{B$_{0.05}$} and \textbf{B$_{0.01}$}. These values correspond to the lowest BPCER achievable while maintaining an APCER of $10\%$ and $5\%$, respectively. In an ideal scenario, a MAD algorithm deployed in a real-world setting would aim to achieve a low APCER (allowing only a minimal number of criminals to bypass detection) of around $0.1\%$, while simultaneously maintaining an acceptable BPCER (generating few false positives) of approximately $5\%$~\cite{ferrara2022morph}.

\begin{table*}[t!]
    \centering
    \begin{tabular}{l|c|ccc|ccc|ccc}
        \toprule
        \multirow{2}{*}{\textbf{Method}} & \textbf{Year} & \multicolumn{3}{c|}{\textbf{Accomplice}} & \multicolumn{3}{c|}{\textbf{Criminal}} & \multicolumn{3}{c}{\textbf{Both}} \\
         & & \textbf{EER} & \textbf{B$_{0.05}$} & \textbf{B$_{0.01}$} & \textbf{EER} & \textbf{B$_{0.05}$} & \textbf{B$_{0.01}$} & \textbf{EER} & \textbf{B$_{0.05}$} & \textbf{B$_{0.01}$} \\
         \midrule
        Demorphing \cite{ferrara2017face} & 2017 & .160 & .377 & .618 & .027 & .015 & .050 & .111 & .235 & .522 \\
        SVM+LBP \cite{scherhag2018towards} & 2018 & .200 & .327 & .623 & .185 & .335 & .697 & .192 & .330 & .628 \\
        DFR \cite{scherhag2020deep} & 2020 & .178 & .482 & .770 & .068 & .090 & .295 & .128 & .347 & .690 \\
        Siamese \cite{borghi2021double} & 2021 & .153 & .347 & .563 & .061 & .095 & .370 & .115 & .257 & .515 \\
        MagFace \cite{kessler2023towards} & 2023 & .112 & .230 & .465 & .028 & .013 & .052 & .083 & .135 & .393 \\
        \midrule 
        \textbf{ACIdA (Ours)} & 2023 & \textbf{.102} & \textbf{.192} & \textbf{.333} & \textbf{.023} & \textbf{.010} & \textbf{.027} & \textbf{.070} & \textbf{.105} & \textbf{.280} \\
        \bottomrule
    \end{tabular}
    \caption{Morphing detection scores obtained on the FEI Morph test dataset across different Differential MAD (D-MAD) competitors presented in the literature. 
    Results are reported in terms of Equal Error Rate (EER), the lowest BPCER related to APCER $\leq 1\%$ and $\leq 5\%$, respectively (see Sect.~\ref{sec:metrics}).
    As reported, the proposed method achieves the best accuracy across all the competitors both in the ``accomplice'' and in the ``criminal'' benchmarks. The generalization capabilities are confirmed in the ``both'' case, in which our model effectively handles different attempt typologies.}
    \label{tab:sota}
\end{table*}

\begin{figure}[h!]
    \centering
    \includegraphics[width=0.5\linewidth]{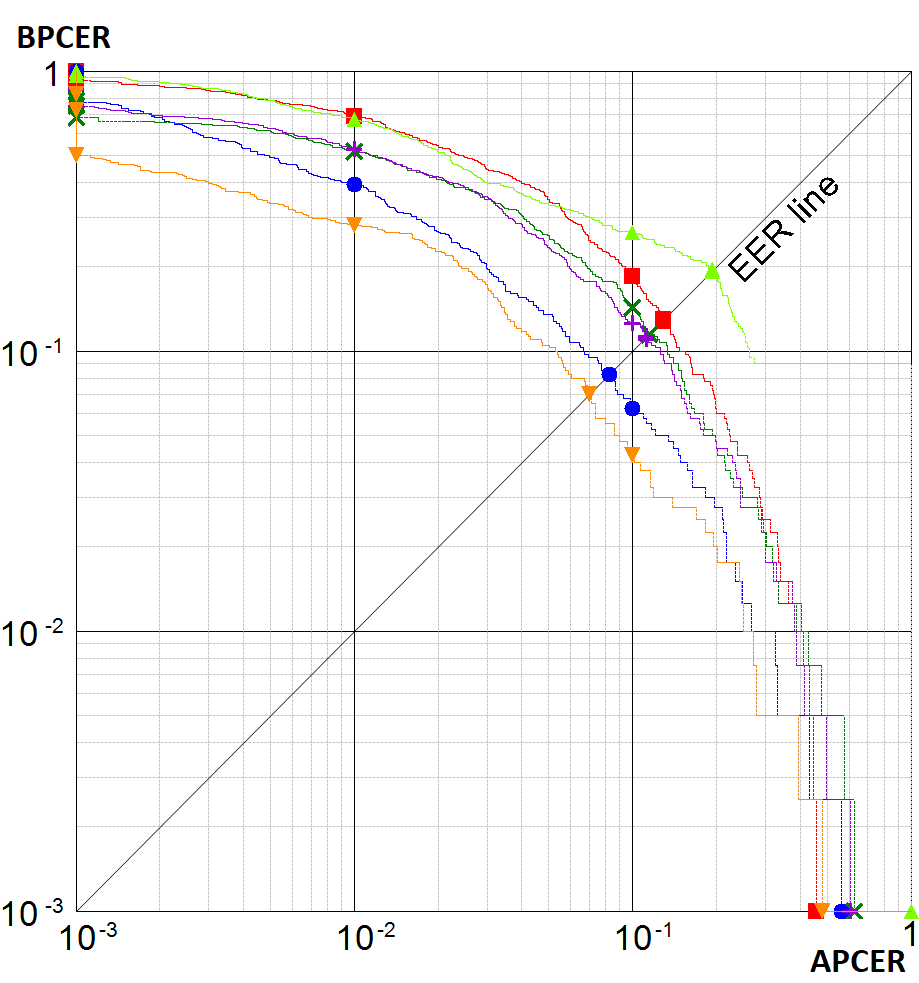}
    \caption{Detection Error Trade-off (DET) curves computed on the FEI Morph dataset considering several literature competitors.
    Competitor reported: ACIdA in orange, \cite{meng2021magface} in blue, \cite{deng2019arcface} in red, \cite{borghi2021double} in dark green, \cite{scherhag2018towards} in light green and \cite{ferrara2017face} in purple. [Better in color.]}
    \label{fig:det}
\end{figure}

The Equal Error Rate (EER) is a commonly reported metric, representing the error rate at which the BPCER and APCER are equal. It is typically presented as a single value, providing a summary measure of the system's performance. 
Besides, APCER and BPCER metrics can be condensed in the Detection Error Trade-off (DET) curve, reported as well to improve the understanding and the comparison of the experimental analysis.

In this paper, we introduce and exploit also the metric Weighted Average Error Metric (WAE) in order to summarize all the performance indicators in a single value (as done in Figure \ref{fig:mad_vs_id}). This metric is formally defined as:
\begin{equation}
    \text{WAE} = w_E E^T
\end{equation}
where $E$ is the set of error-based metric values 
$E = $[ \text{EER}, \textbf{B$_{0.1}$}, \textbf{B$_{0.05}$}, \textbf{B$_{0.01}$}] 
and $w_E = [0.3,\, 0.1, \, 0.2,\, 0.4]$.
These weights are chosen by assigning the majority of the weight to the most common real-world operating point (\textit{i.e.} \textbf{B$_{0.01}$}), followed by the EER, as it is useful for evaluating the performance of the system at a glance, and finally the other two chosen operating points (\textit{i.e.} \textbf{B$_{0.05}$} and \textbf{B$_{0.1}$}).

\subsection{Training procedure}
In the IdA module, the MLP has an architecture composed of $3$ hidden layers of size $250$, $125$, and $64$, with ReLU activation and a single output neuron with a sigmoid function.
For the training, we adopt the Adam~\cite{kingma2014adam} optimizer with an initial learning rate of $10^{-5}$.
For the training of the Inception-ResNet, we adopt the SGD optimizer with a learning rate of $10^{-3}$ and an early-stopping procedure (patience of 5 epochs with a minimum improvement of $10^{-4}$). No momentum decay is exploited.

In the other modules, SVM classifiers share the same details: they implement the Radial Basis Function (RBF) kernel, and are trained with the regularization parameter $C = 1.0$, and the kernel coefficient $\gamma = 10^{-3}$. 

\begin{table*}[th!]
    \centering
    \begin{tabular}{r|ccc|ccc|ccc}
        \toprule
        \multirow{2}{*}{\textbf{Feat. Embed.}}  & \multicolumn{3}{c|}{\textbf{Accomplice}} & \multicolumn{3}{c|}{\textbf{Criminal}} & \multicolumn{3}{c}{\textbf{Both}} \\
         & \textbf{EER} & \textbf{B$_{0.05}$} & \textbf{B$_{0.01}$} & \textbf{EER} & \textbf{B$_{0.05}$} & \textbf{B$_{0.01}$} & \textbf{EER} & \textbf{B$_{0.05}$} & \textbf{B$_{0.01}$} \\
        \midrule
        ArcFace~\cite{deng2019arcface}  & \underline{.115} & \underline{.255} & \underline{.507} & \underline{.056} & \underline{.063} & \underline{.115} & \underline{.093} & \underline{.142} & \underline{.385} \\
        DLib~\cite{king2009dlib}  & .176 & .375 & .587 & .103 & .155 & .240 & .148 & .255 & .517 \\
        SFace~\cite{zhong2021sface}  & .213 & .480 & .640 & .102 & .167 & .327 & .166 & .363 & .585 \\
        Facenet~\cite{schroff2015facenet}  & .129 & .308 & .510 & .061 & .075 & .142 & .097 & .175 & .435 \\
        MagFace~\cite{meng2021magface}  & \textbf{.102} & \textbf{.192} & \textbf{.333} & \textbf{.023} &  \textbf{.010} & \textbf{.027} & \textbf{.070} & \textbf{.105} & \textbf{.280} \\
        \bottomrule
    \end{tabular}
    \caption{Morphing detection scores obtained by the proposed system on the FEI Morph dataset using different feature embeddings originally developed for the Face Recognition task.
    In bold the best results, underlined the second ones.}
    \label{tab:features_comparison}
\end{table*}

\begin{table*}[th!]
    \centering
    \begin{tabular}{cc|ccc|ccc|ccc}
        \toprule
         \multicolumn{2}{c|}{\textbf{Module}} & \multicolumn{3}{c|}{\textbf{Accomplice}} & \multicolumn{3}{c|}{\textbf{Criminal}} & \multicolumn{3}{c}{\textbf{Both}} \\
         \textbf{IdA} & \textbf{Id} & \textbf{EER} & \textbf{B$_{0.05}$} & \textbf{B$_{0.01}$} & \textbf{EER} & \textbf{B$_{0.05}$} & \textbf{B$_{0.01}$} & \textbf{EER} & \textbf{B$_{0.05}$} & \textbf{B$_{0.01}$} \\
        \midrule
         \ding{51} & \ding{51} & \textit{.102} & \textit{.192} & \textit{.333} & \textit{.023} &  \textit{.010} & \textit{.027} & \textit{.070} & \textit{.105} & \textit{.280} \\
         \midrule
         \ding{51} &  & .120 & \textbf{.207} & \textbf{.455} & .117 & .195 & .415 & .120 & .203 & .440 \\
        \ding{51}$^*$ & & .125 & .230 & .480 & .125 & .230 & .480 & .125 & .230 & .480 \\
          & \ding{51} & \textbf{.112} & .230 & .465 & \textbf{.028} & \textbf{.013} & \textbf{.052} & \textbf{.083 }& \textbf{.135} & \textbf{.393} \\
        \bottomrule 
        \addlinespace[1ex] \multicolumn{11}{l}{\small$^*$ Only the Inception-ResNet architecture of IdA module is tested (see Fig.~\ref{fig:general}).}
    \end{tabular}
    \caption{Ablation analysis results obtained on the FEI Morph dataset. 
    As reported, the single modules of the proposed system -- IdA (Sect.~\ref{sec:Ida_module}) and Id (Sect.~\ref{sec:id_module}) -- are separately tested. In addition, we compute the results obtained using only the Inception-ResNet architecture of the IdA module.
    }
    \label{tab:ablation}
\end{table*}

\subsection{Results} \label{sec:results}
The results of the proposed MAD method compared with the literature are reported in Table~\ref{tab:sota}, while the DET curve is reported in Figure~\ref{fig:det}. 
As shown, our method is able to overcome the competitors in all the investigated benchmarks. 
In particular, our method not only achieves better performance in comparison with D-MAD based only on identity comparison~\cite{scherhag2020deep,kessler2023towards,ferrara2017face} but also with methods that include the analysis of artifacts introduced by the morphing procedure on input images~\cite{borghi2021double,scherhag2018towards}.
We note also that, in all cases, the error rates obtained in the accomplice scenario are higher, confirming that the proposed scenario introduces specific challenges and requires specific investigations in future D-MAD works.
It is important to note that our method achieves great accuracy in the ``criminal'' benchmarks, confirming its effectiveness also in a more traditional D-MAD working scenario.
Finally, the results of the ``both'' case, suggest that the inclusion of an Attempt classification module, \textit{i.e.} a mechanism that gives the possibility to apply different solutions with criminals and accomplices, is a key element in improving overall performance, as also investigated in the following.

\begin{table*}[th!]
\centering
    \begin{tabular}{c|cc}
    \toprule
    \textbf{Classifier} & \textbf{Accuracy} & \textbf{F1-score}  \\
    \midrule
    \textbf{SVM}                & \textbf{0.650}             & \textbf{0.634} \\
    \midrule
    \textbf{Random forest}      & 0.575             & 0.554  \\
    \textbf{AdaBoost}           & 0.603             & 0.580  \\
    \textbf{KNN} (K=5)                & 0.575             & 0.545  \\
    \textbf{Decision Tree}      & 0.572             & 0.554 \\
    \textbf{MLP}                & 0.639                  &  \textbf{0.634} \\
    \bottomrule 
\end{tabular}
\caption{Comparison of several classifiers of the Attempt Classification (AC) module, using as input feature embeddings extracted using the method described in \cite{meng2021magface}.}
\label{tab:classifier_comparison}
\end{table*}

\begin{table*}[th!]
    \centering
    % \resizebox{1\linewidth}{!}{
    \begin{tabular}{ccc|ccc|ccc|ccc}
        \toprule
        \multicolumn{3}{c}{\textbf{Classification Attempt (CA)}} & & \multicolumn{8}{c}{\textbf{Morphing Attack Detection (IdA - Id)}}\\
        \multirow{2}{*}{\textbf{Classifier}} & \multirow{2}{*}{\textbf{Accuracy}}  & \multirow{2}{*}{\textbf{BF to}} & \multicolumn{3}{c|}{\textbf{Accomplice}} & \multicolumn{3}{c|}{\textbf{Criminal}} & \multicolumn{3}{c}{\textbf{Both}} \\
         & & & \textbf{EER} & \textbf{B$_{0.05}$} & \textbf{B$_{0.01}$} & \textbf{EER} & \textbf{B$_{0.05}$} & \textbf{B$_{0.01}$} & \textbf{EER} & \textbf{B$_{0.05}$} & \textbf{B$_{0.01}$} \\
        \midrule
        \multirow{2}{*}{Oracle} & \multirow{2}{*}{100\%} &  IdA & \textit{.120} & \textit{.207} & \textit{.455} & \textit{.208} & \textit{.522} & \textit{.625} & \textit{.146} & \textit{.480} & \textit{.598} \\
        & & Id & \textit{.006} & \textit{.002} & \textit{.005} & \textit{.028} & \textit{.013} & \textit{.052} & \textit{.018} & \textit{.005} & \textit{.040} \\
        \midrule
        \multirow{2}{*}{SVM} &\multirow{2}{*}{66.9\%} & IdA & .171 & .495 & .560 & .046 & .027 & .555 & .118 & .445 & .555 \\
          & & Id  & \textbf{.102} & \textbf{.192} & \textbf{.333} & \textbf{.023} & \textbf{.010} & \textbf{.027} & \textbf{.070} & \textbf{.105} & \textbf{.280} \\
        \bottomrule
    \end{tabular}
    % }
    \caption{Classification performance and morphing detection scores obtained on the FEI Morph dataset. In particular, we highlight the impact of the attempt classification accuracy on the whole proposed system, with respect to the performance of an ``oracle'' classifier reported on the top lines.}
    \label{tab:third}
\end{table*}

In the second part of our analysis, we test how the performance of the proposed system is influenced when two different score fusion techniques are employed to create the final MAD score, namely ``weighted'' and ``selection'':
the former produces the final morphing score by summing the ones produced by the different modules, each weighted by the probabilities returned by the Attempt Classification module, as described in Equation~\ref{eq:weighted} and depicted in Figure~\ref{fig:general};
the latter directly returns the score produced by the module whose associated probability computed by the Attempt Classification module is the highest, as shown in Equation~\ref{eq:selection}, where $p_{max} = \max\left(p_A, p_B, p_C\right)$.

\begin{equation}
    S = \begin{cases}
        S_{Id} & \text{if } p_B = p_{max} \wedge p_C = p_{max} \\
        S_{IdA} & \text{if } p_C = p_{max} \\
    \end{cases}
    \label{eq:selection}
\end{equation}

Experimental results show that the weighted strategy is overall more effective than employing the selection fusion technique, respectively totaling on the global test set $\text{EER} = 0.070$ versus $0.076$, $\text{B}_{0.05} = 0.105$ versus $0.125$, and $\text{B}_{0.01} = 0.280$ versus $0.385$.
Therefore, the weighted sum fusion strategy is adopted in our framework.

In Table~\ref{tab:features_comparison}, we include an analysis of the feature embedding source: indeed, several works on the Face Recognition task have been recently introduced in the literature~\cite{schroff2015facenet,deng2019arcface,meng2021magface,zhong2021sface}, constantly improving the accuracy, and can be exploited in our proposed MAD pipeline.
In particular, we focus our analysis on recent and state-of-the-art methods. 
According to the analysis provided by the Deepface framework~\footnote{\url{https://github.com/serengil/deepface}}, we select best-performing algorithms on the Labeled Faces in the Wild~\cite{huang2008labeled} dataset. 
In addition, we include in our analysis also the recently introduced and promising MagFace~\cite{meng2021magface} approach.
It is important to note that one of the top-performing methods, ArcFace~\cite{deng2019arcface}, has already been effectively utilized in the field of D-MAD~\cite{scherhag2020deep}.
As reported, MagFace achieves the best performance, confirming that superior accuracy in the Face Recognition task leads to better performance also in the MAD task, especially if based on the identity comparison.

In Table~\ref{tab:ablation}, the ablation study of the proposed method is reported.
In particular, we test the performance of the system using only the IdA or the Id module and exploiting the previously determined best feature (\textit{i.e.} MagFace~\cite{meng2021magface}). 
In addition, we report the results exploiting only the Inception-Resnet architecture, which computes the morphing score relying only on the document image, in a S-MAD fashion.
Experimental results confirm that the adoption of both modules is essential to obtain good performance, and that, as expected, the second module, \textit{i.e.} Id, is particularly able to detect the morphed images in the ``criminal'' benchmark, while the IdA module obtains interesting results, in terms of BPCERs (\textbf{B$_{0.05}$}, \textbf{B$_{0.01}$}), in the accomplice one.

In conclusion of our experimental analysis, we focus our investigation on the Attempt Classification (AC) module. 

In the first experiment, reported in Table \ref{tab:classifier_comparison}, we test a variety of different classifiers. 
Interestingly, experimental results reveal that the SVM is the best choice, also with respect to deep learning-based solutions (\textit{i.e.} MLP). 

In addition, we analyze the impact of the classification accuracy of the adopted SVM classifier on the whole MAD pipeline.
Indeed, in order to have an upper bound result, we replace in the AC module, the SVM classifier with an oracle, \textit{i.e.} a classifier able to perfectly predict the classes -- criminal accomplice and bona fide -- of the input attempts. 
In addition, we analyze how the overall performance of the system is affected when the morphing score of bona fide pairs is computed using $S_{IdA}$ instead of $S_{Id}$ (see Eq.~\ref{eq:weighted}).

Results are reported in Table~\ref{tab:third}.

Firstly, the accuracy of the SVM classifier with respect to the oracle reveals that there is a margin for improvement in the attempt classification task.
Closing this gap can significantly enhance the overall performance, as demonstrated by the impressive MAD results reported in the oracle case, with an $\text{EER} = 0.006$.

Secondly, it is possible to note that a substantial improvement in the overall MAD performance is achieved by properly selecting the module used in the weighted average score computation for bona fide images.
Specifically, the use of the Id module with bona fide images represents the best solution, since analyzing artifacts produced by the morphing procedure is not useful -- bona fide images are free of visible or not visible artifacts -- and it could even be counterproductive.
Indeed, in the MAD literature, the S-MAD task, and specifically the artifact detection task, is more challenging with respect to the D-MAD one~\cite{raja2020morphing}.

Finally, we test the proposed system on the FVC-onGoing platform. 
It is worth noting that publicly available platforms \cite{cit:nist,fvc_ongoing} do not provide an effective test bed for the scenario described in this paper, since typically the accomplice attempts are not considered when the presence in the morphed image is unbalanced in favor of the accomplice.
Results are reported in Table~\ref{tab:fvc_comparison}: we observe that the proposed method is able to achieve good accuracy in any case, in particular with respect to the previous work~\cite{di2023combining}.
In this criminal-based scenario, DFR~\cite{scherhag2020deep} remains the preferable option to use to achieve good performance, since it is based on highly effective identity features when the comparison involves only the criminal or the morphed images equally represents the two contributing subjects (\textit{i.e.} the morphing factor is $0.5$).

\begin{table*}[th!]
\centering
\begin{tabular}{rcccc}
\textbf{Algorithm}                   & \textbf{EER}              & \textbf{B}$_{0.1}$             & \textbf{B}$_{0.05}$            & \textbf{B}$_{0.01}$            \\
\toprule
\textbf{DFR} \cite{scherhag2020deep}                        & 4.54                      & 2.00                      & 3.93                      & 18.87                     \\
\textbf{Demorphing} \cite{ferrara2017face}                  & 14.17                     & 17.20                     & 22.77                     & 65.57                     \\
\textbf{Siamese} \cite{borghi2021double} & \multicolumn{1}{l}{23.37} & \multicolumn{1}{l}{35.03} & \multicolumn{1}{l}{48.97} & \multicolumn{1}{l}{93.60} \\
\textbf{MBLBP} \cite{scherhag2017vulnerability}   & \multicolumn{1}{l}{33.47} & \multicolumn{1}{l}{52.80} & \multicolumn{1}{l}{59.93} & \multicolumn{1}{l}{74.80}\\
\textbf{WL} \cite{damer2019detecting}                  & 37.13                      & 71.67                     & 83.27                      & 95.67                     \\
\textbf{BSIF} \cite{kannala2012bsif}                  & 45.93                     & 78.30                     & 84.13                     & 93.83                     \\
\textbf{DN} \cite{singh2019robust}                  & 52.03                     & 89.70                     & 94.70                     & 98.57                     \\
\textbf{Laplace} \cite{kraetzer2017modeling}                  & 55.13                     & 96.70                     & 98.67                     & 99.87                     \\
\midrule
\textbf{R-DMAD} \cite{di2023combining}                      & 10.23                     & 10.33                     & 19.67                     & 47.47                     \\
\textbf{ACIdA}                       & 7.84                      & 7.57                      & 12.60                     & 26.23                     \\
\bottomrule
\end{tabular}
\caption{Comparison of the results on the sequestered DMAD-SOTAMD\_D-1.0 benchmark through the FVConGoing platform~\cite{fvc_ongoing}. As shown, the proposed method is able to generalize on a different scenario, obtaining a good accuracy w.r.t. the other D-MAD methods available in the literature.}
\label{tab:fvc_comparison}
\end{table*}

\section{Conclusion}
ACIdA, a novel modular D-MAD method was proposed, and extensive experimental validation demonstrated its ability to achieve state-of-the-art results, surpassing competing methods in the proposed scenario. These results suggest that an analysis of possible morphing artifacts in combination with the use of identity features can increase the robustness of D-MAD approaches when dealing with a high subject similarity. 
Furthermore, a new scenario was introduced, expanding the scope of application of current MAD systems based on two input images (D-MAD), by comparing trusted live acquisition images with both criminal and accomplice subjects. Our preliminary analysis indicates that this new scenario is challenging and raises new interesting research aspects in the D-MAD field, especially for identity-based systems.
The findings of this study contribute to the advancement of MAD systems, enhancing the security and reliability of FRS in the face of morphing attacks. Further research can build upon these results to develop even more robust and efficient MAD systems in the future.
In particular, new experimentation and research are needed to improve the classification accuracy of the attempt that, as shown, is a key element for the final performance of the proposed system. 
Finally, we suggest the possibility of integrating the proposed novel scenario in the two main web platforms with sequestered datasets for the MAD task, \textit{i.e.} NIST FATE MORPH and FVC-onGoing, that, at the time of writing, offers D-MAD tests with only the criminal comparison.

%Bibliography
\bibliographystyle{unsrt}  
\bibliography{main.bib}

\end{document}